\documentclass[11pt,a4paper]{article}
\usepackage[hyperref]{eacl2021}
\usepackage{times}
\usepackage{latexsym}

\usepackage{microtype}

\aclfinalcopy % Uncomment this line for the final submission
\usepackage{caption}
\usepackage{subcaption}
\usepackage{amssymb}

\usepackage{siunitx} % To align the numbers later on
\usepackage{bbm} % indicator function
\usepackage{booktabs}
\newcommand{\ra}[1]{\renewcommand{\arraystretch}{#1}}
\usepackage{multirow, makecell}
\usepackage{rotating}
\usepackage[utf8]{inputenc}

\usepackage{etoolbox}
\usepackage{tikz}
\usepackage{fontawesome}
\usepackage[ruled,vlined]{algorithm2e}
\usepackage{tabularx}

\newcolumntype{K}[1]{>{\centering\arraybackslash}p{#1}}

\usepackage{url}
\usepackage{amsmath}
\usepackage{float}
\restylefloat{table}
\newcommand\Mark[1]{\textsuperscript#1}

\title{A Neural Few-Shot Text Classification Reality Check}
\author{
    Thomas Dopierre\Mark{1}\Mark{2} \and Christophe Gravier\Mark{1} \and \\\and \textbf{Wilfried Logerais\Mark{2}}
\\ \\
\begin{tabular}{*{2}{>{\centering}p{.4\textwidth}}}
\Mark{1}Laboratoire Hubert Curien\\UMR CNRS 5516\\Université Jean Monnet&   \Mark{2}Meetic \\Paris, France \tabularnewline 
Saint-Étienne, France &  \tabularnewline
\url{firstname.lastname@univ-st-etienne.fr} & \url{{t.dopierre,w.logerais}@meetic-corp.com}
\end{tabular}
}

\date{}

\begin{document}
    \maketitle
    \begin{abstract}
        Modern classification models tend to struggle when the amount of annotated data is scarce. To overcome this issue,
        several neural few-shot classification models have emerged, yielding significant progress over time, both in Computer Vision and Natural Language Processing. %
        In the latter, such models used to rely on fixed word embeddings before the advent of transformers. Additionally, some models used in Computer Vision are yet to be tested in NLP applications. In this paper, we compare all these models, first adapting those made in the field of image processing to NLP, and second providing them access to transformers. We then test these models equipped with the same transformer-based encoder on the intent detection task, known for having a large number of classes. %As a reality check, we conduct a study providing transformers to the most popular neural few-shot solutions for a fair comparison, applied on several intent detection datasets. %We also conduct another set of experiments on three different public Intent Detection datasets, where the number of labels is one order of magnitude higher, and sentences are way shorter.
        Our results reveal that while methods perform almost equally on the ARSC dataset, this is not the case for the Intent Detection task, where the most recent and supposedly best competitors
        %(which had access to transformers when devised)
        perform worse than older and simpler ones 
        (while all are given access to transformers). We also show that a simple baseline is surprisingly strong.
        All the new developed models, as well as the evaluation framework, are made publicly
        available\footnote{https://github.com/tdopierre/FewShotText}.
    \end{abstract}

    \section{Introduction}\label{sec:introduction}

% motivations for few shots
    Text classification often requires a large number of mappings between texts and target classes, so that it is challenging to build few-shot text classification models~\cite{induction}. With the recent advances of transformer-based models~\cite{devlin2018bert,Wolf2019HuggingFacesTS}
    along with their fine-tuning techniques~\cite{sun2019fine}, text classification has significantly improved. %, although it seems mainly thanks to an upgrade of the text feature extraction. %Consequently, the glass ceiling of these improvements is made of the clever Hans effect~\cite{} (such new state-of-the-art may not be meaningful~\cite{Niven} since transformer-based models mainly learn from statistical cues present in their training and fine-tuning datasets).
% and domain variability (depending on the domain, extracted text representations can still be hard to separate) and
    In few-shot settings, methods based on these extracted text representations have been historically made of semi-supervision, especially thanks to pseudo-labeling~\cite{blum1998combining,NLP:self-training,tri-training},
    which aims at propagating known
    labels to unlabeled data points in the representational space. Such methods depend on the number of collected unlabeled data,
    which can also be costly to obtain~\cite{Charoenphakdee2019}, and also suffer from the infamous pipeline effect in NLP~\cite{tenney2019},
    as cascade processing tends to make errors accumulate. In order to address the hindrance of collecting unlabeled data,
    modern approaches include
    unsupervised data augmentation techniques~\cite{uda}. It consists of generating samples through well-established text
    augmentation techniques in Neural Machine Translation, such as backtranslation~\cite{sennrich2015improving,edunov2018understanding},
    and then use a consistency
    loss, training the classifier to assign the same prediction to all variations of the same sample text.
    While collecting new pseudo-labels can therefore be overcome by manipulating the dataset (especially using data augmentation techniques),
    the pipelining error accumulation effect instead calls for new neural architectures supporting scarcity of labeled data in
    an end-to-end fashion.
    %As image processing few-shot models suffered from the same issue with the ImageNet models moment earlier than we are with the transformer moment in NLP, it is only natural to the NLP community to see how image
    %classification models came with end-to-end solutions.
    % matching : inventé pour des images, relation : images, induction : meta-learning for texts, protopp: images, proto:images
    Such end-to-end few-shot neural architectures for few-shot classification were discovered in image processing -- it includes Matching Networks~\cite{vinyals2016matching}, Prototypical Networks~\cite{snell2017prototypical} plus a follow-up known as Prototypical Networks++~\cite{protopp}, and Relation Networks~\cite{Sung_2018_CVPR}.
    Ultimately Induction Networks~\cite{induction} is a meta-learning based method dedicated to few-shot text classification, supposedly the state-of-the-art. Since our contribution considers this family of models, we will further detail them in Section~\ref{sec:algorithms}.
    Nonetheless, it is important to stress that most of these neural architectures were originally devised to integrate image feature extractors. Despite both text and image relying on features extractors, a paragraph or sentence of few words hardly
    convey as much information as a full-fledged three-canals $600\times 400$ image ($720,000$ numerical values intrinsically).
    It is therefore of the utmost practical interest to validate and compare if what works best for end-to-end few-shot \textit{image}
    classification is the same for end-to-end few-shot \textit{text} classification. Moreover, when applying these end-to-end few-shot models to text, two main system components are into action: the text feature extractor itself and the downstream part of the neural network that provides a learning strategy over few shots. If we want
    to compare these systems, we need to plug the same feature extractor (hopefully the best one, that is transformer-based currently)
    into each end-to-end model. For the time being, the literature on end-to-end few-shot text classification compare
    aforementioned techniques using a different text extractor for each system, which is the one available when the technique was discovered -- these text encoding varying greatly (Section~\ref{subsec:encoder}). From that point-of-view, it is hardly possible to conclude if the improvement over time in few-shot text classification is due to new few-shot learning techniques or plainly to the significant advances made by text feature extractors. The same applies to vectors metrics: one method can use the cosine and another the euclidean distance, and that choice alone can impact conclusions made on the method being the state-of-the-art, although it could well rely only on the metric at work. Ultimately, experimental setups are usually restricted to one dataset, and evaluation schemes are heterogeneous among papers~\cite{Yu2018}, % (binary classification for ARSC~\cite{}, C-way K-shots for other datasets~\cite{}, hiding or not hiding some classes at training time~\cite{}, setting the same shots for each  episode\footnote{https://github.com/Gorov/DiverseFewShot\_Amazon}, etc.) making it even harder to measure the progress in the field. In that context, our contributions are summarized as follows:
    \begin{itemize}
        \item We revise different end-to-end neural architectures for few-shot text classification using the \textit{same} transformer-based feature extractor,
        \item We investigate how these re-implemented state-of-the-art solutions compete with very simple baselines found to be yet very competitive for few-shot classification in the field of image-processing,
        \item We introduce an evaluation framework based on a number of intent detection datasets which is significantly bigger than what is usually used as evaluation in seminal papers transposing each of these architectures from image to text classification,
        \item The entire framework used in this paper, including all the re-implemented methods plugged with up-to-date transformers, is provided as an open-source repository for further research.
    \end{itemize}

    In a nutshell, we will demonstrate that providing a transformer-based encoder to a previously obsolete few-shot technique
    makes it the state-of-the-art again, that standard baselines are surprisingly strong, and that Induction Networks,
    while performing well for binary sentiment classification, struggles to perform correctly in the most common setups
    of few-shot text classification.

    %    \begin{itemize}
%        \item Induction network (sota) reality check: full transformer pour tous le monde (remise des choses à plat / corriger l'écart d'extraction de features)
%        \begin{itemize}
%            \item Classif binaire vs k-way y-classes
%            \item Phrases positives/négatives
%        \end{itemize}
%        \begin{itemize}
%            \item Quel pont/différence avec l'application des ces méthodes à l'image (alors que la mode d'extraction de features est différent)
%        \end{itemize}
%        \item Faire un benchmark des méthodes neuronales E2E few shots pour Intent Detection
%    \end{itemize}

    %  \section{Related Works}\label{sec:related-works}
    %  \textbf{related-works}

    \section{Few-Shot Classification Methods}\label{sec:algorithms}

    \begin{figure*}

        \begin{subfigure}{.5\textwidth}
            \centering
            % include first image
            \includegraphics[width=1\linewidth]{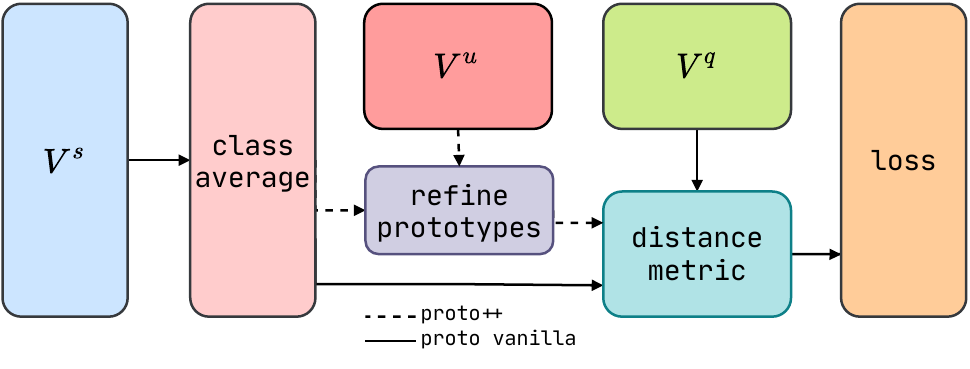}
            \caption{Prototypical Network, with the optional proto++ step. In the original Prototypical Networks, the euclidean distance
            is used as distance metric.}
            \label{fig:sub-proto}
        \end{subfigure}
        \hskip 0.5cm
        \begin{subfigure}{.5\textwidth}
            \centering
            % include second image
            \includegraphics[width=.9\linewidth]{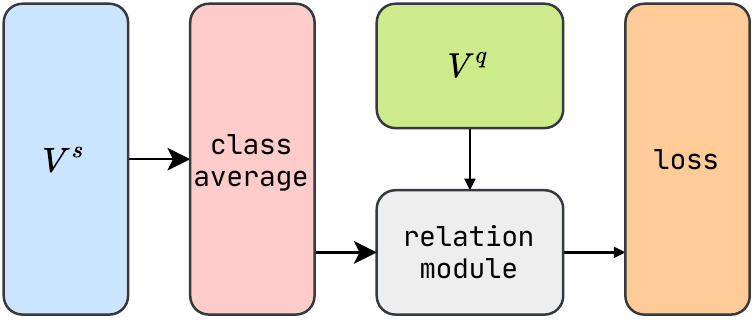}
            \caption{Relation Network}
            \label{fig:sub-relation}
        \end{subfigure}
        \vspace{0.2cm}
        \begin{subfigure}{.45\textwidth}
%                \centering
            % include third image
            \includegraphics[width=\linewidth]{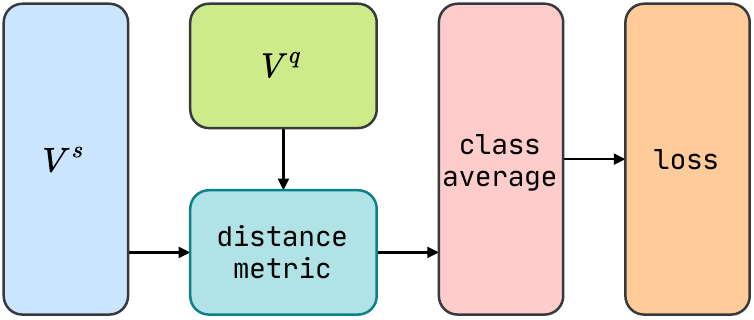}
            \caption{Matching Network. In the original Matching Networks, the cosine similarity is used as a distance metric.}
            \label{fig:sub-matching}
        \end{subfigure}
        \hskip 1.7cm
        \begin{subfigure}{.45\textwidth}
            \centering
            % include fourth image
            \includegraphics[width=\linewidth]{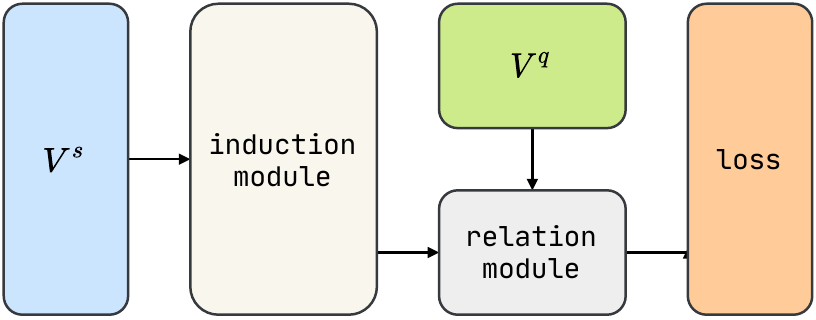}
            \caption{Induction Network}
            \label{fig:sub-induction}
        \end{subfigure}
        \caption{Few-Shot classification methods and variants used in our experiments.}
        \label{fig:fewshotsmethods}
    \end{figure*}

    In this section, we will describe the few-shot learning methods.
    In the following section, sentence vectors derived from the sentence encoder are denoted $v$.
    $V^s$, $V^q$, and $V^u$ represent vectors for support, query, and unlabeled points, respectively.
    The number of shots is denoted $K$, and the number of classes per episode is denoted $C$.
    The k$^{th}$ support vector of class $c$ is denoted $v^{s}_{c, k}$.
    In the equations, $s^{q}_{i,j}$ (resp. $s^{q}_{i, c}$) will denote the similarity between the i$^{th}$ query vector
    and the j$^{th}$ support vector (resp. the c$^{th}$ class). Similarly, $s^{u}_{i, j}$ represents the similarity between the i$^{th}$
    unlabeled sample and the j$^{th}$ support vector.
    When needed, the number of unlabeled data is denoted $U$.
    For each method relying on
    a given similarity or distance metric, we devise two experiments, using either the cosine similarity or the euclidean distance. Those additional experiments are crucial, as they allow us to compare methods directly, without introducing a metric choice bias.
    Architectures of the different few-shot approaches are illustrated in Figure~\ref{fig:fewshotsmethods}.
    They are each detailed in Section~\ref{subsec:matching-networks} and onwards, yet we first introduce the common building blocks
    among all methods in what follows.

    \subsection{Common building blocks}\label{subsec:common-building-blocks}

    \paragraph*{Class average}
    All Matching, Prototypical, and Relation Networks contain a \textit{class average} block. This step is used to directly compare a query point to a given class in order to make a prediction for this query point. In both Matching and Prototypical Networks,
    this step averages embeddings of support points for each class (they are class \texttt{prototypes}), which are then compared to query points to output class probabilities. In Matching Networks, this block averages similarity scores class-wise.
    On the contrary, in Induction Networks, this step is lacking as support points
    are converted into prototypes using an Induction Layer, which aims at finding a better way to aggregate such knowledge than using the average (Section~\ref{subsec:induction-networks}).

%    In both approaches, this block intervenes at a different step, which will be developed in their corresponding
    \paragraph*{Loss}
    Matching and Induction Networks both use the mean squared error (MSE) loss. Other methods use cross-entropy (CE). We implemented both losses on Matching and Induction networks, and it leads to very similar results -- and sometimes, slightly better using CE. We therefore report results for all models using CE due to space limitations. Note that both losses are available in the publicly available source code.
    The cosine similarity being bounded, it would not make sense to directly apply such a loss on cosine similarities. To overcome this issue, we multiply the cosine similarities by a constant factor of $5$, allowing them to reach more extreme values, hence ensuring that probabilities obtained by softmax are sparse enough.

%    \hl{Rque: Fig1 contient cosine et euclidean blocks, sauf que dans le texte a la section au dessus on dit jsutement
%    qu'on va tester les deux pour toutes les methods où cela s'applique... Peut être changer le text des ces block cosine
%    et euclidiean distance en un meme nom identique ? De la meme facon que relation module apparait pour relation et
%    induction, on pourrait avoir un vector metric qui apparait pour proto et matching, sans les décrire dans ce paragraph
%    d'intro des trucs en communs, de la meme facon qu'on en decrit pas relation module comme pas commun au quatre approches}

    \subsection{Matching Networks}\label{subsec:matching-networks}
    Introduced by~\citet{vinyals2016matching}, Matching Networks (Figure~\ref{fig:sub-matching}) rely on the comparison between query and support vectors using the cosine similarity in the seminal paper.
    After similarities between a query point and all support points are computed, they are averaged for each class. The predicted label for a given query point is the one with the highest average cosine similarity. In our notation framework, this process is summed up in Equation~\ref{eq:matching-network}.
    \begin{equation}
        \label{eq:matching-network}
        s^{q, \text{matching}}_{i,c} = K^{-1}\sum_{k=1}^{K}\frac{\left(v^{q}_{i}\right)^{T}v^{s}_{c,k}}{\|v^{q}_{i}\|_{_{2}}\|v^{s}_{c,k}\|_{_2}}
    \end{equation}

    \subsection{Prototypical Networks}\label{subsec:prototypical-networks}
    Prototypical Networks (Figure ~\ref{fig:sub-proto}) were introduced by ~\citet{snell2017prototypical} as an extension of Matching Networks. After obtaining support vectors from the encoder,
    a class-wise average operation is done, as in Equation~\ref{eq:prototypes}. This results in $C$ prototypes denoted
    $\left\{p_c, c \in [\![1, C ]\!]\right\}$, each one being the representative of a class.
    Then, a distance metric compares all query points to all prototypes. For each query point, the predicted class is the one for which this distance is the smallest. In the original Prototypical Networks,
    the euclidean distance was used, as in Equation~\ref{eq:prototypical-network}.
    We also add the cosine similarity-based distance in our experiments in order to measure the impact
    of selecting another distance metric.

    \begin{equation}
        \label{eq:prototypes}
        p_{c} = K^{-1}\sum_{k=1}^{K}v^{s}_{c,k}
    \end{equation}

    \begin{equation}
        \label{eq:prototypical-network}
        s^{q,\text{proto}}_{i, c} = \frac
        {\exp\left(-\|v^{q}_{i} - p_{c}\|^{2}_{2}\right)}
        {\sum_{c'=1}^{C}\exp\left(-\|v^{q}_{i} - p_{c'}\|^{2}_{2}\right)}
    \end{equation}

    An extension to Prototypical Networks was proposed by~\citet{protopp}, where unlabeled data points are used along with support and query points. After computing each class's prototype, a soft k-means technique is applied to further refine those prototypes using unlabeled data points. The refined prototypes, denoted $\tilde{p}_c$, are derived using Equation~\ref{eq:prototypespp}. This additional step aims at correcting the support points selection bias and making the method more robust.

    \begin{equation}
        \label{eq:prototypespp}
        \tilde{p}_{c} = \frac
        {\sum_{k=1}^{K}v^{s}_{c, k} + \sum_{i=1}^{U}v^{u}_{i}s^{u, \text{proto}}_{i, c}}
        {K + \sum_{i}s^{u, \text{proto}}_{i, c}}
    \end{equation}

    \subsection{Relation Networks}\label{subsec:relation-networks}
    Relation Networks~\cite{Sung_2018_CVPR} challenge the idea of using a pre-defined metric. % attention a la difference distance vs metric
    The Relation Module takes as an input a query vector $v^{q}_{i} \in \mathbb{R}^d$, and the prototype of a class $p_c \in \mathbb{R}^d$, the latter being obtained the same way as in Prototypical Networks (Equation~\ref{eq:prototypes}). The idea is to use a relation module, modeling the relationship between those
    two vectors, yielding a
    similarity score $s_{i,c} \in \left(0, 1\right)$. Instead of using a pre-defined distance metric like the euclidean or the cosine one,
    this approach allows such networks to learn this metric by themselves. Two different relation module architectures exist.

    \paragraph{base} The \texttt{base} relation module concatenates both $v^{q}_{i}$ and $p_c$, and applies a small feed-forward neural network composed of
    two linear layers, with a ReLU activation function in between.
    The formula for this given relation module is described in Equation~\ref{eq:base-relation-module}, where $C\left(\cdot, \cdot\right)$ denotes
    the concatenation operator, $f\left(\cdot\right)$ denotes the ReLU activation function, and $w, M_1, M_2$ are learnable parameters.

    \begin{equation}
        \label{eq:base-relation-module}
        s^{q, \text{rel-base}}_{i,c} = \left< w,M_2\left(f\left(M_1\left(C\left(v^{q}_{i}, p_c\right)\right)\right)\right)\right>
    \end{equation}

    \paragraph{NTL} Introduced by ~\citet{ntl}, the \textbf{N}eural \textbf{T}ensor \textbf{L}ayer
    relation module uses intermediate learnable matrices $M_k \in \mathbb{R}^{d, d}$ to model the relation between support vectors and prototypes.
    The similarity score for this relation module is obtained using Equation~\ref{eq:ntl-relation-module-1}, where $w$ is a learnable parameter.
    Following the work done by ~\citet{induction}, we fix the number $h$ of intermediate matrices to 100 in all our experiments.

    \begin{equation}
        \label{eq:ntl-relation-module-1}
        s^{q, \text{rel-ntl}}_{i,c} = \left< w, z^{\text{rel-ntl}}_{i,c} \right> \hskip .3cm , \hskip .3cm w \in \mathbb{R}^{h}
    \end{equation}
    \begin{equation}
        \label{eq:ntl-relation-module-2}
        z^{q, \text{rel-ntl}}_{i,c,t} = f\left(\left(v^{q}_{i}\right)^T M_t p_{c}\right) \hskip .3cm , \hskip .3cm t \in [\![ 1, h ]\!]
    \end{equation}

    \subsection{Induction Networks}\label{subsec:induction-networks}

    Induction Networks~\cite{induction} aims at finding a general representation of each class in the support set to compare to new queries.
    They are composed of both an induction module and a relation module.
    The main motivation for such networks is that representing the class by the average vector of its data
    points -- what is done in Prototypical and Relation networks -- is too restrictive.
    The first part, the induction module, leverages a dynamic routing~\cite{sabour2017dynamic} algorithm. In their contribution,
    \citet{induction} show that their method can better \textit{induce} (hence their name) and generalize class-wise
    representations. For the second part, an NTL Relation Module is used: this is the same as the one introduced earlier (Section~\ref{subsec:relation-networks}). Such networks are illustrated in Figure~\ref{fig:sub-induction}.

    As in~\cite{induction}, we fix the number of routing iterations to $3$, and the number of matrices in the NTL to $100$.

    \subsection{Classifier Baselines}\label{subsec:classifier}

    Few-shot learning algorithms are designed to overcome the data scarcity problem. With the tremendous shift in the architecture of sentence encoders using transformers, control baselines are needed to validate their ability to learn from few samples. For this reason, we include as a first \texttt{Baseline} model a traditional classifier,
    as described by Equation~\ref{eq:baseline}, added on top of BERT. Both $W$ and $b$ are learnable parameters,
    fine-tuned on the support vectors $V^s$. In our experiments, this method will henceforth be referred to as \texttt{Baseline}.

    \begin{equation}
        \label{eq:baseline}
        s^{q, \text{baseline}}_{i, j} = \left( W v^{q}_{i} + b\right)_{j} \hskip .3cm ; \hskip .3cm W \in \mathbb{R}^{C, d}
    \end{equation}

    In addition to this \texttt{Baseline} model, we also implement a variant of it, which will henceforth be referred to as \texttt{Baseline++}. In that second baseline, the classifier design differs as follows: it measures similarities to a learnable vector instead of transforming vectors into logits using a linear layer.
    The matrix $W$ used in the \texttt{Baseline} model can be writen as $\left[w_1, \ldots, w_C\right]$ where each $w_k \in \mathbb{R}^d$ is a weight vector
    corresponding to the $k^{th}$ class. To measure the similarity between class $j$ and a query vector $v^{q}_{i}$, we compute the similarity scores in Equation~\ref{eq:baselinepp}.
    After all scores $s_{\cdot, \cdot}$ are computed, we then obtain a probability vector through normalization using the softmax function).

    \begin{equation}
        \label{eq:baselinepp}
        s^{q,\text{baseline++}}_{i,j} = \frac{w_j^{T}v^{q}_{i}}{\|w_j\|_{_2}\|v^{q}_{i}\|_{_2}}
    \end{equation}

\begin{figure}
        \begin{subfigure}{\linewidth}
            \centering
            % include first image
            \includegraphics[width=.9\linewidth]{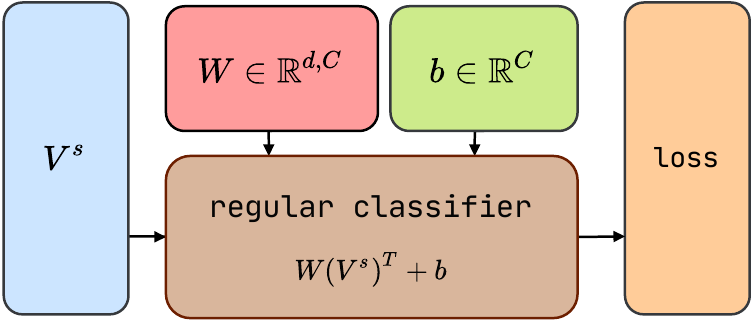}
            \caption{Baseline Network.}
            \label{fig:sub-baseline}
        \end{subfigure}
        \newline
        \vskip 5px
        \begin{subfigure}{\linewidth}
            \centering
            % include third image
            \includegraphics[width=.9\linewidth]{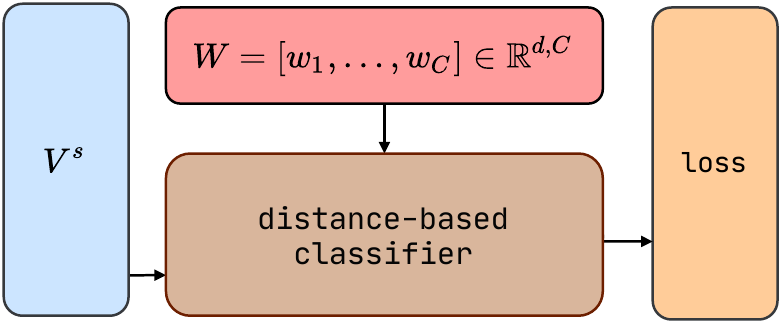}
            \caption{Baseline++ Network. The distance-base classifier can either use cosine or euclidean distance}
            \label{fig:sub-baselinepp}
        \end{subfigure}
        \caption{Few-Shot classification baselines used in our experiments.}
        \label{fig:baselines}
    \end{figure}

    As in Prototypical Networks, the derived vectors $\left[w_1, \ldots, w_C\right]$ can be interpreted as class prototypes.
    For both baselines, at each training episode, the weights $W$ and $b$
    are initialized, and the whole model is fine-tuned for a few iterations using support samples.
    This is important in practice, as it teaches the sentence encoder -- a transformer, see Section~\ref{subsec:encoder} -- how to produce good enough embeddings for the downstream classifier to learn efficiently. At test time, the same process is used -- using test labels --, except that we freeze the encoder's weights and only fine-tune the classifier part. The baselines architectures are represented in Figure~\ref{fig:baselines}.

    \section{Experimental Setup}\label{sec:datasets}

    \subsection{Few-Shot Evaluation Setup}\label{subsec:few-shot-evaluation-setup}
    Introduced by~\citet{vinyals2016matching}, few-shot classification corresponds to the case when a classifier must adapt to new classes, denoted here as $\mathcal{C}_{test}$, unseen during training, and only given a few labeled examples of these new classes. To this end, the approaches assume that during training, a task-significant set of classes noted $\mathcal{C}_{train}$ is available, along with an accordingly task-significant number of labeled data for each class $c_{train_{i}}\in \mathcal{C}_{train}$. For each training episode, $C$ classes are sampled from $\mathcal{C}_{train}, C \ll |\mathcal{C}_{train}|$. Then, $K$ support examples and
    $Q$ query examples are randomly drawn for each of these classes. The model is then iteratively trained using both query and support points.

    At testing time, the same sampling strategy is made, this time drawing classes among $\mathcal{C}_{test}$, with $\mathcal{C}_{test} \cap \mathcal{C}_{train} = \emptyset$. The model is then evaluated on its ability to predict labels for the $Q$ query samples, using the $K$ support samples (unless otherwise stated, $C$, $Q$, and $K$ values are the same at both testing and training time).

    This training procedure is called $C$-way $K$-shot classification. In all our experiments, we used $K=Q=5$. Concerning the value of $C$, it is fixed
    to $2$ for ARSC, as this dataset is already composed of binary classification tasks.
    Regarding the intent detection datasets we introduce later (Section~\ref{subsec:datasets}),
    in order to see the shift between ARSC binary tasks and the more common $5$-way evaluation~\cite{induction,protopp}, we measured
    performances of the different models with $C$ ranging from $2$ to $5$.

    \subsection{Datasets}\label{subsec:datasets}
    In this section, we describe the datasets used in our evaluation framework. The first one is a popular sentiment classification dataset,
    while the others are intent detection datasets. All datasets are public and in English.

    \paragraph{ARSC} The \textbf{A}mazon \textbf{R}eview \textbf{S}entiment \textbf{C}lassification dataset~\cite{ARSC} is composed of product reviews
    from $27$ product categories. Each review belongs to one of the $27$ domains, and contains a grade ranging from $1$ to $5$ stars.
    The usual setup~\cite{yu2018diverse,induction} to evaluate few-shot classification with this dataset is as follows:
    for each of $p \leq 27$ product category and $2\leq t \leq 5$ score thresholds, $\mathcal{E}_{ARSC} = p \times t$ binary
    classification evaluation tasks are created. In each of these $p\times 4$ tasks, a competitor model must learn to
    classify negative ($< t$) and positive ($\geq t$) reviews. To build our test tasks, we consider the same product categories
    as previous works ~\citep{yu2018diverse, induction}, which are \textit{Books}, \textit{DVD}, \textit{Electronics}, \textit{Kitchen},
    and $t=3$ (thresholds are picked in the $\{2, 4, 5\}$ set) -- hence 12 binary classification test tasks in our benchmark for this dataset.

    Each of these twelve evaluation tasks comes with a number of support test samples ($K=Q=5$ as stated previously).
    Nonetheless, in \citep{yu2018diverse} the same $5$ samples per testing class are fixed for all
    experiments\footnote{See labeled sampled in \url{https://github.com/Gorov/DiverseFewShot\_Amazon}},
    which leads to a significant selection bias towards these $5$ randomly selected samples used throughout the evaluation.
    In order to get more consistent results, we ran additional experimental runs, each of them selecting randomly new support samples. In the ARSC result table (table~\ref{tab:arsc-results}), this corresponds to the last column
    (\texttt{BERT + Sample shots}).

    \paragraph{OOS} The \textbf{O}ut \textbf{O}f \textbf{S}cope dataset\footnote{\url{https://github.com/clinc/oos-eval}}~\cite{OOS} is an intent detection dataset containing $150$ equally-distributed classes.
    While initially used for out-of-scope prediction, it was also motivated by a high number of classes, a low number of examples per class ($150$), and its chatbot life-like style. In our experiments, we discard the out-of-scope class, keeping the remaining $150$ classes to work with.

    \paragraph{Liu} Introduced by~\citet{Liu-dataset}, this intent detection dataset consists in $54$ classes. This dataset was collected
    on the Amazon Mechanical Turk platform, where workers were given an intent and had to formulate queries for this intent with their own words.
    It is highly imbalanced: the most common class (\texttt{query}) holds $5,920$ samples while the least common one (\texttt{volume\_other}) $24$  samples.

    \paragraph{TREC28} TREC\footnote{\url{https://trec.nist.gov/data/qa.html}} is an open-domain fact-based dataset for question classification. %Because such questions could be asked to a chatbot, we associate this dataset to the intent classification field.
    We use the $50$ labels version of the dataset but remove the labels which have less than $40$ samples.
    This filtering process yields a dataset with $28$ classes, ranging from $40$ to $962$ samples per class.

    \subsection{Sentence Encoder}\label{subsec:encoder}
%\hl{Dire le nom du réseau en plus de la ref}
    In previous works comparing few-shot text classification methods, sentence encoders were not always the same. For example, ~\citet{yu2018diverse}
    use a CNN on top of word embeddings, while ~\citet{induction} use a Bi-LSTM. Those differences make the results hard to compare since they do not use the same method to convert sentences into vectors. In our experiments, in order to reduce this selection bias, and since
    it is now the state-of-the-art in many applications, we use a BERT~\cite{devlin2018bert} encoder, using models from the Hugging Face~\cite{Wolf2019HuggingFacesTS}
    team.

    \begin{table*}
        \ra{0.8}
        \centering
        \resizebox*{\textwidth}{!}{
            \begin{tabular}{@{\hskip 1em}K{3.8cm}@{\hskip 1em}K{1cm}@{\hskip 1em}K{1.8cm}@{\hskip 1em}K{3cm}@{\hskip 1em}K{4cm}@{\hskip 1em}K{2.1cm}}
                \toprule
                & \multicolumn{2}{c}{Configuration} & \multicolumn{3}{c}{Mean binary accuracy}\\
                \cmidrule(r{0.5em}){2-3}
                \cmidrule(l{0.5em}){4-6}
                Model & {Metric} & {Relation module} & {Original encoder~$\dagger$} & {BERT as encoder $(\nearrow$ or $\searrow$ w.r.t. original encoder)} & {BERT + Sample shots}\\
                \midrule \\[-1em]
                \multirow{2}{*}{\shortstack[1]{Matching Network\\ \cite{vinyals2016matching}}} & euclid. & N/A & $-$ & 81.2 & 82.9 \\
                & cosine & N/A & 65.7 & 81.9 ($\nearrow$) & 83.3 \\[6pt]

                \multirow{2}{*}{\shortstack[1]{Prototypical Network\\ \cite{snell2017prototypical}}} & euclid. & N/A & 68.2 & 80.0 ($\nearrow$) & 82.6 \\
                & cosine & N/A & $-$ & 81.7 & 83.5 \\[6pt]

                \multirow{2}{*}{\shortstack[1]{Proto++\\ \cite{protopp}}} & euclid. & N/A & \faCameraRetro & 82.4 & \textbf{84.0} \\
                & cosine & N/A & \faCameraRetro & \textbf{82.6} & 83.6 \\[6pt]

                \multirow{2}{*}{\shortstack[1]{Relation Network\\ \cite{Sung_2018_CVPR}}} & N/A & base & $-$ & 81.0 & 82.9 \\
                & N/A & ntl & 83.1 & 81.7 ($\searrow$) & 83.3\\[6pt]

                {\shortstack[1]{Induction Network\\ \cite{induction}}} & N/A & ntl & 85.6 & 79.3 ($\searrow$) & 80.3 \\[6pt]

                {Baseline} & N/A & N/A & \faCameraRetro & 80.7 & 79.8 \\[6pt]

                \multirow{2}{*}{Baseline++} & euclid. & N/A & $-$ & 81.9 & 82.2 \\
                & cosine & N/A & \faCameraRetro & 79.7 & 81.1 \\

                \bottomrule
            \end{tabular}
        }
        \caption{Mean accuracy on the 12 ARSC binary classification test tasks. In column~$\dagger$, results are reproduced from the Induction Networks seminal
        paper~\cite{induction} (where applies), a dash ($-$) means that results for that encoder/metric pair were not reported,
        and \faCameraRetro~denotes models only tested on computer vision tasks (first time applied to text in our contribution). The \textit{BERT} column is our implementation using the same 5 shots as the first column but using a BERT encoder for all methods. The last column is also using BERT, but results are averaged over five runs, sampling different shots for each run. In the Configuration column, N/A means that the configuration criteria does not apply to the model.}
        \label{tab:arsc-results}
    \end{table*}
    For each dataset, instead of using an off-the-shelf pre-trained model, we fine-tune it on the masked language modeling task, as it greatly improves the quality of embeddings~\cite{sun2019fine,uda}. This fine-tuned transformer is then used as input for all few-shot models.

    \section{Observations}\label{sec:observations}

    We report results for the ARSC dataset in Table~\ref{tab:arsc-results}, and results for the Intent Detection
    tasks in Table~\ref{tab:intents-results}.

    \subsection{Baselines are surprisingly strong}\label{subsec:baselines-strong}
    Few-shot learning methods were originally used to overcome data scarcity. In those situations, training a classifier on top of a small dataset -- in our case, 5 samples per class -- can be hard. However, our experiments on ARSC show that the \texttt{Baseline} and \texttt{Baseline++}, plain and simple classifiers, get surprisingly close to state-of-the-art results. Table~\ref{tab:baseline-examples} provides four correct and four incorrect classification examples for the \texttt{Baseline} model.
    
    \begin{table}[h!]
        \ra{0.85}
        \resizebox{\columnwidth}{!}{%
            \centering

            \begin{tabular}{ll}
                \toprule
                \multicolumn{2}{c}{Correct classification examples}\\
                \midrule
                S: & Do I have enough in my boa account for a new pair of skis ?\\
                P: & {\textbf{\texttt{balance}}} \\
                T: & {\textbf{\texttt{balance}}} \\
                \midrule
                S: & What's 15\% of 68 ?\\
                P: & {\textbf{\texttt{calculator}}} \\
                T: & {\textbf{\texttt{calculator}}} \\
                \midrule
                S: & I need to know the nearest bank's location.\\
                P: & {\textbf{\texttt{directions}}} \\
                T: & {\textbf{\texttt{directions}}} \\
                \midrule
                S: & What do I take home ?\\
                P: & {\textbf{\texttt{income}}} \\
                T: & {\textbf{\texttt{income}}} \\[10pt]
                \toprule
                \multicolumn{2}{c}{Incorrect classification examples}\\
                \midrule
                S: & On Tuesday you are supposed to have a meeting.\\
                P: & {\textbf{\texttt{meeting\_schedule}}} \\
                T: & {\textbf{\texttt{calendar}}} \\
                \midrule
                S: & What are my insurance rewards ? \\
                P: & {\textbf{\texttt{insurance}}} \\
                T: & {\textbf{\texttt{redeem\_rewards}}} \\
                \midrule
                S: & How much farther is Orlando from my location? \\
                P: & {\textbf{\texttt{current\_location}}} \\
                T: & {\textbf{\texttt{distance}}} \\
                \midrule
                S: & Stop talking please. \\
                P: & {\textbf{\texttt{change\_speed}}} \\
                T: & {\textbf{\texttt{cancel}}} \\
                \bottomrule
            \end{tabular}
            \caption{Examples of OOS query examples correctly and incorrectly predicted by the \texttt{Baseline} method using 5 shots. $S$ (resp. $P$, $T$) is the sentence (resp. prediction and true label).}
            \label{tab:baseline-examples}
        }
    \end{table}
    While it fails to predict the correct text label for some shots, it is also able to correctly classify sentences
    such as \textit{What do I take home ?} among the $50$ test classes of the OOS dataset. On the ARSC dataset, it is also important to note
    that the \texttt{Baseline++} model is significantly better than the \texttt{Baseline}, and is even on par with
    all other architectures, except \texttt{Prototypical Networks}.
    %With the recent discovery of very large language models that have been shown by~\cite{GPT3}, where a transformer succeeds in learning from few shots.

    \subsection{Sample selection bias}\label{subsec:sample-selection-bias}
    The mean accuracy difference between the last and the second columns of Table~\ref{tab:arsc-results} accounts for the difference of randomly selecting new support samples at each iteration (last column) as opposed to picking the same fixed pool of support samples as done previously (second to last column). We can see that this difference alone is in the range of the increments brought by each model over time (baselines aside, bringing from $1$ point up to $2.6$ points for Prototypical Networks). This huge gap shows the importance of using evaluation tricks like cross-validation,
    instead of evaluating only for one run over a fixed set of shots.

    \subsection{Impact of switching to transformers}\label{subsec:switching-to-transformers}
    One of the main contributions of our paper is to compare few-shot learning methods with the lowest bias possible (see Section~\ref{subsec:encoder}).
    On the ARSC dataset, using transformers drastically changes the performances of all methods. When feeding the same transformer-based encoder to all few-shot methods, Prototypical Networks are now on
    top, whereas metric learning approaches (Induction \& Relation Networks) tend to struggle, almost reaching the same performances as Matching Networks.
    \begin{table*}
        \ra{0.9}
        \centering
        \resizebox*{\textwidth}{!}{
            \begin{tabular}{ccc @{\hskip 1em}  cccc @{\hskip 2em} cccc @{\hskip2em} cccc}
                \toprule
                & \multirow{3}{*}{ Metric}
                & \multirow{3}{*}{
                    \small
                    \begin{tabular}{c}
                        Relation \\[4px]
                        Module
                    \end{tabular}
                }
                & \multicolumn{4}{c@{\hskip 2em}}{Liu}
                & \multicolumn{4}{c@{\hskip 2em}}{OOS}
                & \multicolumn{4}{c}{TREC28}
                \\
                \cmidrule(lr{2.5em}){4-7}
                \cmidrule(lr{2em}){8-11}
                \cmidrule(lr{1em}){12-15}
%                \\
                & &
                & 2 & 3 & 4 & 5
                & 2 & 3 & 4 & 5
                & 2 & 3 & 4 & 5 \\
                \midrule
                \multirow{2}{*}{Matching} & euclid. & -
                & 96.6 & 93.7 & 91.1 & 89.1
                & 99.2 & 98.7 & 98.1 & 97.7
                & 89.4 & 81.6 & 76.6 & 69.6\\

                & cosine & -
                & 93.3 & 87.9 & 84.8 & 81.0
                & 96.8 & 95.8 & 95.1 & 94.7
                & 81.6 & 75.4 & 68.5 & 63.5\\[5px]

                \multirow{2}{*}{Proto} & euclid. & -
                & 97.4 & 95.3 & 93.4 & 91.8
                & \textbf{99.5} & 99.0 & 98.7 & 98.4
                & \textbf{92.6} & \textbf{87.6} & 82.0 & \textbf{79.2}\\

                & cosine & -
                & 94.6 & 90.4 & 88.5 & 85.6
                & 97.6 & 97.3 & 96.9 & 96.5
                & 85.6 & 79.1 & 74.5 & 71.3\\[5px]

                \multirow{2}{*}{Proto++}
                & euclid. & -
                & \textbf{97.7} & \textbf{95.7} & \textbf{93.7} & \textbf{92.2}
                & 99.5 & \textbf{99.1} & \textbf{98.8} & \textbf{98.5}
                & 91.7 & 84.9 & \textbf{82.0} & 76.8\\

                & cosine & -
                & 94.0 & 90.9 & 87.9 & 85.4
                & 97.5 & 97.3 & 97.0 & 96.5
                & 83.8 & 78.1 & 71.0 & 65.9\\[5px]

                \multirow{2}{*}{Relation}
                & - & base
                & 88.2 & 76.5 & 71.8 & 65.1
                & 91.1 & 86.0 & 79.9 & 77.9
                & 80.8 & 66.3 & 61.7 & 51.8\\

                & - & ntl
                & 87.4 & 80.1 & 74.3 & 69.0
                & 90.9 & 84.2 & 82.0 & 77.8
                & 74.7 & 62.5 & 57.7 & 48.6 \\[5px]

                {Induction} & - & ntl
                & 73.9 & 57.9 & 52.6 & 40.6
                & 74.9 & 59.3 & 50.9 & 43.8
                & 70.3 & 49.6 & 41.9 & 33.9\\[5px]

                {Baseline} & - & -
                & 94.3 & 89.0 & 84.1 & 79.8
                & 99.1 & 98.5 & 97.7 & 97.2
                & 90.5 & 83.6 & 79.3 & 75.7\\[5px]

                \multirow{2}{*}{Baseline++} & euclid. & -
                & 93.1 & 87.6 & 81.4 & 78.1
                & 95.8 & 93.3 & 92.1 & 90.6
                & 87.7 & 78.3 & 72.5 & 69.1\\

                & cosine & -
                & 93.1 & 86.8 & 81.0 & 75.1
                & 98.9 & 97.9 & 96.8 & 96.1
                & 86.7 & 78.2 & 72.1 & 70.0\\

                \bottomrule
            \end{tabular}
        }
        \caption{Mean accuracy of $C$-way $5$-shot intent detection, with $C$ ranging between $2$ and $5$.
        Each reported value is the average over five runs with different random seeds.
        For each column, the best method is highlighted in \textbf{bold}.}
        \label{tab:intents-results}
    \end{table*}
    Such metric learning approaches rely on various weight matrices and parameters, while more traditional approaches (\texttt{Matching}, \texttt{Proto})
    do not use any parameter apart from the encoding step. This hints that the upstream transformer does most of the learning and is able to model the embedding space well enough such that no more additional metric learning is needed.
    The massive increase in embedding quality brought by the BERT encoder makes Prototypical Network approaches
    reclaim the state-of-the-art position.
%    with a , while keeping the original \texttt{Proto} method or its variant (\texttt{Proto++}) intact. %In practice, metric learning-based methods, which are longer to train, may not be needed to provide state-of-the-art performances. % genre on est polis xD

%    \begin{itemize}
%        \item Différence sur ARSC avant / après
%        \item commentaires associés
%    \end{itemize}

    \subsection{The curious case of induction networks}\label{subsec:results-induction}
    When ~\citet{induction} introduced Induction Networks, both the ARSC dataset and a private intent detection dataset were used for evaluation (publicly unavailable).
    Our experiments of this method on the ARSC dataset confirm those results in an acceptable range, even when trying to get more consistent results using multiple random seeds. Nonetheless, the performances of this method are underwhelming on all three intent detection datasets, even when matching the binary classification scenario using $C=2$. Those poor performances were observed
    both on the test set as well as the train set, discarding the over-fitting argument.
    Such a big performance gap between sentiment and intent classification tasks show that Induction Networks,
    while suited for the former, are not directly applicable to any type of task.

%    \begin{itemize}
%        \item Cas d'évaluation classif binaire dans leur papier (et pourquoi ils font cela)
%        \item XPs en plus sur la façon d'évaluer classif binaire (pour s'assurer que c'est robuste)
%        \item pourtant dire que pas adapter au cas plus générale de K-way C-classes dixit nos résultats
%    \end{itemize}

    \subsection{On metric choice}\label{subsec:on-metric-choice}
    Prototypical Networks were originally designed to do better than Matching Networks. The two differences between them are
    the placement of the \texttt{class average} step, and the choice of the metric (cosine for Matching, euclidean for Prototypical).
    Our results show that metric choice yields a big gap in performances for both methods, this gap being larger than the gap caused by the model design. This hints that when using a pre-defined metric -- excluding the case of metric learning --, choosing the right metric is of paramount importance. Moreover, while \texttt{Matching Networks} were designed to use the cosine distance, we found here that they perform significantly better when equipped with the Euclidean distance (on all datasets for all number of given test classes).

    \subsection{On architectural choices}\label{subsec:on-architectural-choices}
    Overall, Prototypical Networks come on top of every intent detection dataset. More importantly, their gap between other competing approaches is wider as the number of classes increases. This result is important, as in practice, the number of classes is likely to be higher
    than what is used in the literature. 
    %In such a case, having a model that scales well with the number of classes is crucial. 
    The extended variant, proto++,
    obtains mixed results. %, as it sometimes yields small improvements over the vanilla version (Liu, OOS), and sometimes significantly degrades performances (TREC28).
    While this shows that using unlabeled data can have some benefits, we also observe that the proto++ way of integrating this external knowledge is perfectible. Ultimately, note that our results do not mirror Computer Vision results.
    Since few-shot learning methods are used on top of embeddings, we could emit the hypothesis that they can be applied to any embeddings,
    regardless of the field. However, while Relation Networks, for example, were performing well in Computer Vision classification tasks
    -- the tasks which they were originally designed for -- as well as text classification -- back in the days when transformers did not exist --,
    this is not the case anymore. The drawback is that all methods are very sensitive to the feature extractor used in prior steps.

%    \begin{itemize}
%        \item Proto/matching : le choix de la métrique importe plus que le choix de l'archi matching ou proto
%        \item Proto++ tient mieux l'augmentation du nombre de classes
%        \item Différence par rapport à résultat en image
%    \end{itemize}

    \section{Conclusion}\label{sec:conclusion}
    We provided a fair comparison for end-to-end neural few-shot text classification methods discovered over the last few years. When they are all equipped with a transformer-based text encoder,
    we show that Prototypical Networks become the state-of-the-art again. We also found that a traditional classifier trained on few shots yields very competitive results, especially when given shots are re-sampled at each iteration. Ultimately, we also demonstrated the significant impact of the vector metric, illustrated by Matching Networks strongly improving by only replacing the cosine by the euclidean distance. The complete source code with the re-implementation of all the tested methods and evaluation framework used in this study is publicly
    available\footnote{https://github.com/tdopierre/FewShotText} -- we hope that it will help the community build upon consistent comparative experiments. % and foster few-shot text end-to-end text classification.

%    \section*{Acknowledgments}
%    Hidden during the review process.
    %The authors would like to thank Professor Pierre Maret from laboratoie Hubert Curien UMR CNRS 5516 at Université Jean Monnet for his very constructive comments. We also gratefully acknowledge the support of the NVIDIA Corporation with the donation of one NVIDIA TITAN X GPU for our research.
    %
    \bibliographystyle{acl_natbib}
    \bibliography{anthology,mybib}

\end{document}